\documentclass[10pt,letterpaper]{article}

\usepackage{cogsci}
\usepackage{graphicx}
\usepackage{caption}
\usepackage{multirow}
\cogscifinalcopy 

\usepackage{cmap}
\usepackage[T1]{fontenc}
\usepackage[american]{babel}
\usepackage{csquotes}
\usepackage{newtxtext,newtxmath}  

\usepackage[
  backend=biber,
  style=apa,
  natbib=true,
  annotation=false,
]{biblatex}
\usepackage{float} 

\usepackage{enumitem}

\usepackage[hidelinks]{hyperref}

\title{Human-Like Anaphor Resolution in Large Language Models}

\newcommand{\gtlogo}{\raisebox{0pt}{\includegraphics[scale=0.04]{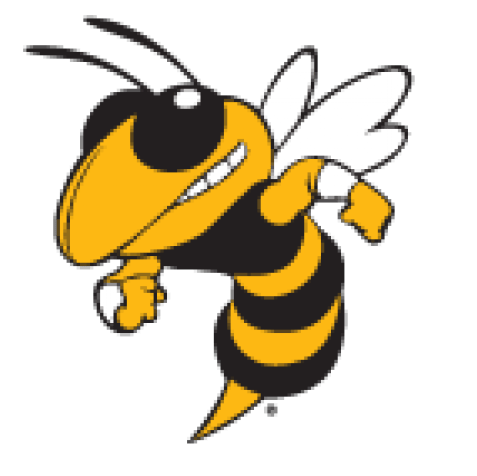}}}
\author{{\large \bf Keane Zhang, Varshini Chinta, Raj Sanjay Shah, Sashank Varma} \\
  \{keane, vchinta6, rajsanjayshah,
varma\}@gatech.edu
  \\
  Georgia Institute of Technology \gtlogo
  }

\begin{document}

\maketitle

\let\thefootnote\relax
\footnotetext{Code for reproducing all model evaluations, analyses, and figures is available at:
\href{https://github.com/wristy/anaphor}{\texttt{github.com/wristy/anaphor}}.}
\setcounter{footnote}{0}

\begin{abstract}

\emph{Anaphors} are expressions that refer to other expressions, called \emph{antecedents}. The process of connecting the two is called \emph{resolution}. 
Cognitive science has identified multiple factors that affect the speed and success of anaphor resolution, including discourse structure, situation-model properties, and semantic factors.
Here, we investigate whether these factors also affect anaphor resolution in five Large Language Models (LLMs) with open weights: GPT-2-XL, Llama-3.1-8B, Pythia-12B, Mistral-7B, and Mistral-24B. 
To model processing difficulty, we adopt the standard linking hypothesis that relates human reading times to model surprisal at the anaphor.
As a second behavioral measure, we compare model accuracy to human accuracy on comprehension questions probing the antecedents of anaphors.
The results show selective cognitive alignment: some LLMs exhibit human-like sensitivity to discourse prominence and distance-based factors in anaphor resolution, while showing weaker or absent sensitivity to semantic interference effects. These findings delimit the conditions under which LLMs approximate human anaphor resolution.


\end{abstract}

\section{Introduction}

Large Language Models (LLMs) are deep neural networks with millions or billions of parameters trained on large text corpora.
Transformer-based architectures, beginning with BERT \citep{devlin2019bert} and GPT-2 \citep{radford2019language} and extending to more recent families such as GPT \citep{openai_gpt5_2025}, Llama \citep{grattafiori2024the}, and Mistral \citep{mistral_small3_2025}, are highly performant on language comprehension tasks \citep{wang2024mmlu}.
Beyond task performance, these models are increasingly evaluated as candidate models of human cognition \citep{ivanova2025how,piantadosi2024why,shah2025the}, and more specifically as models of human language processing \citep{charpentier-etal-2025-findings,li2024incremental}.

Much of the existing work evaluating LLMs as cognitive models has focused on phenomena at the word and sentence levels, examining model behavior in relatively local and decontextualized linguistic settings. This leaves open the question of whether LLMs capture discourse-level processes that unfold over extended text. Anaphor resolution is one such process: it requires maintaining and retrieving information across sentences, integrating linguistic input with a developing situation model, and resolving interference from competing referents. Modeling these demands goes beyond local prediction and instead requires sensitivity to connected text. In sequential language models, shifts in expected word probabilities, typically operationalized as surprisal, have been shown to track human reading times, providing a process-level link between model predictions and human comprehension behavior \citep{wilcox2020on, li2024incremental}.

Despite growing interest in cognitive alignment, comparatively little attention has been paid to text- and discourse-level phenomena. The current study addresses this gap by focusing on anaphor resolution. It focuses on \emph{anaphors}, which are expressions that refer to other expressions, called \emph{antecedents}. Pronouns are a familiar class of anaphors; more generally, language is rife with referential expressions. The process of connecting an anaphor to its antecedent during online comprehension is called \emph{resolution}. This can be easy when an anaphor and its antecedent occur within the same sentence or within a few sentences of each other; in this case, resolution requires only searching working memory \citep{daneman1980individual}. However, when reading longer texts, anaphors can be separated from their antecedents by many sentences. In this case, resolution requires using an anaphor as a cue to memory and attempting to retrieve its antecedent from the reader's situation model, which is the evolving representation of the overall meaning of the text \citep{mcnamara2009misc,vandijk1983strategies}.

Cognitive science research has investigated the factors that affect the speed and accuracy of anaphor resolution for humans. For example, the greater the number of sentences between an anaphor and its antecedent, the slower it is read, presumably because readers must ``search" farther back in their memory for the text. Reading time is one measure; another is the accuracy of comprehension questions. To continue the example, the greater the number of sentences, the less accurate people are when answering a comprehension question about the antecedent, presumably because there is a greater chance that the resolution failed. \textit{Here, we ask whether LLMs are sensitive to the same factors as human comprehenders during anaphor resolution.} To the extent that they do, LLMs gain credibility as candidate models of discourse-level language processing in cognitive science \citep{ivanova2025how, shah2025the}.

\subsection{Cognitive Science Studies of Anaphor Resolution}

The current study focuses on six factors identified by cognitive science as affecting the speed and accuracy of anaphor resolution.

The first factor is a thematic feature of texts: If the antecedent is \textit{topicalized}, then it should feature more prominently in a reader's situation model, and therefore it should be more accessible when an anaphor to it is encountered. In this case, resolution should be relatively fast and accurate. There are various ways to topicalize an antecedent. For example, \citet{obrienej1987antecedent} manipulated whether or not it was focused by the title of the text. The second factor concerns a surface feature of text itself: The greater the \textit{sentential distance} (i.e., number of sentences) between an anaphor and its antecedent, the slower and less accurate resolution is \citep{clark1979in}. \citet{obrienej1987antecedent} orthogonally varied these two factors, antecedent topicality and sentential distance, and Experiment 1 runs LLMs on the materials of this study.

The third and fourth factors concern not surface distance but \textit{contextual} distance. As readers comprehend a text, they build a model of the story world it describes, variously called the situation model, mental model, or world model \citep{mcnamara2009misc,vandijk1983strategies}. The third factor is the \textit{spatial distance} (i.e., perceived physical distance) between the anaphor and antecedent in the reader's situation model \citep{morrow1987accessibility,rinck1995anaphora}. The fourth factor is the \textit{temporal duration} (i.e., perceived elapsed time) between the two \citep{anderson1983the,varma2019the,zwaan1996processing}. The greater the spatial distance and the longer the temporal duration, the slower and less successful anaphor resolution becomes. The following text illustrates a relatively long temporal duration:

\begin{quote}
\textit{Antecedent: The mechanic kicked \underline{the metal chair}.}\\

\textit{Duration (long): It took \textbf{2 hours} to drive back home and return with the critical tool.}\\

\textit{Anaphor: Afterwards, he felt foolish for having kicked the \underline{metal furniture} when he had been frustrated}.\\
\end{quote}

\noindent \citet{anderson1983the} found that readers are relatively slow to read the anaphor and less accurate in answering a follow-up comprehension question about the antecedent:
\begin{quote}

\textit{What piece of \underline{metal furniture} did the mechanic kick out of frustration?}
\end{quote}
compared to a text with a relatively short temporal duration:

\begin{quote}
\textit{Duration (short): It took \textbf{10 minutes} to drive back home and return with the critical tool.}
\end{quote}

The fifth and sixth factors concern the semantics of the situation models that readers build. The \textit{semantic similarity} between the anaphor and antecedent is how much they resemble each other: the greater the similarity, the better the anaphor is as a cue to memory to retrieve the antecedent from the situation model. Returning to the example above, \textit{chair} is a typical member of the \textit{furniture} category \citep{rosch1976basic}. Thus, the two have high semantic similarity, and this will speed resolution. By contrast, if the antecedent had been atypical (e.g., \textit{stool}), this would have slowed resolution \citep{garrod1977interpreting,varma2019the}. Next, consider if a second member of the \textit{furniture} category had been present in the text, one that is not the antecedent:

\begin{quote}
\textit{Distractor: The mechanic walked past the \underline{wooden desk} and headed to the door.}
\end{quote}

\noindent If it is a typical member of the same category, as in the example above, then this causes \textit{semantic interference} in the search for the correct antecedent (i.e., \textit{chair}) in the situation model, and this will slow resolution and undermine its accuracy. By contrast, if the non-antecedent distractor is atypical (e.g., \textit{wooden shelf}), then this should have only a small deleterious effect \citep{corbett1984prenominal,varma2019the}.

\subsection{Anaphor Resolution in LLMs}

Process-level links between language models and human comprehension are often established using surprisal, which has been shown to correlate with human reading times across a range of syntactic and semantic phenomena \citep{hale2001a,levy2008expectation}. While surprisal-based analyses have been widely applied at the word and sentence levels, their use in evaluating discourse-level processes such as anaphor resolution remains limited \citep{wilcox2020on,li2024incremental}. 

In early work, \citet{hu-etal-2020-closer} and \citet{lee-schuster-2022-language} evaluated the ability of early LLMs (e.g., BERT, GPT-2) to resolve reflexive pronominal references to antecedents within the same sentence, with mixed results even for this simple case. More promisingly, \citet{pandit-hou-2021-probing} evaluated BERT's ability to resolve anaphors over increasing function of sentential distance. They identified attention heads at higher layers that bridge between anaphors and their antecedents, though these connections were weaker at longer distances ($6-10$ sentences). They also used a cloze procedure to probe, at the anaphor position, the model's best guess of the antecedent, finding low accuracy even at close sentential distances (i.e., $\geq 3$ sentences).

Recent work has examined the ability of LLMs to resolve anaphors and coreference relations, either by prompting them directly or by adapting them to existing benchmarks. These studies show substantial variability across models, datasets, and prompt formulations, and general-purpose LLMs often underperform specialized coreference systems \citep{cambria2025semantics, liu2025enhancing}. Moreover, reported gains can be sensitive to dataset artifacts, lexical heuristics, and recency biases, raising questions about whether high accuracy reflects robust discourse understanding or shallow statistical cues \citep{novk2025findings,gan2023assessing}.

More importantly, high coreference accuracy does not imply human-like processing. Existing evaluation practices rely on inconsistent metrics and lack a standard framework for assessing LLM-generated responses, particularly across datasets and task formulations \citep{talukdar2025coreference}. Standard NLP benchmarks do not manipulate or isolate the factors that cognitive science has shown to govern anaphor resolution, such as discourse prominence, distance-based accessibility, semantic similarity, or interference from competing referents. As a result, existing evaluations cannot distinguish between models that resolve anaphors using surface heuristics and those that exhibit sensitivity to the cognitive constraints shaping human comprehension \citep{ivanova2025how,shah2025the}.


\emph{Rather than proposing a new benchmark or optimizing coreference performance, the present study adopts classic cognitive science paradigms as the evaluation framework. We ask whether LLMs exhibit sensitivity to the same discourse, situational, and semantic factors that shape human anaphor resolution, using surprisal and comprehension accuracy as complementary behavioral proxies.}

This framing allows us to evaluate cognitive alignment beyond correct referent identification, focusing instead on whether models are sensitive to the same processing factors that affect human readers.


\begin{figure}[t]
\centering
\includegraphics[trim={0cm 0cm 0cm 0cm}, width=0.5\textwidth]{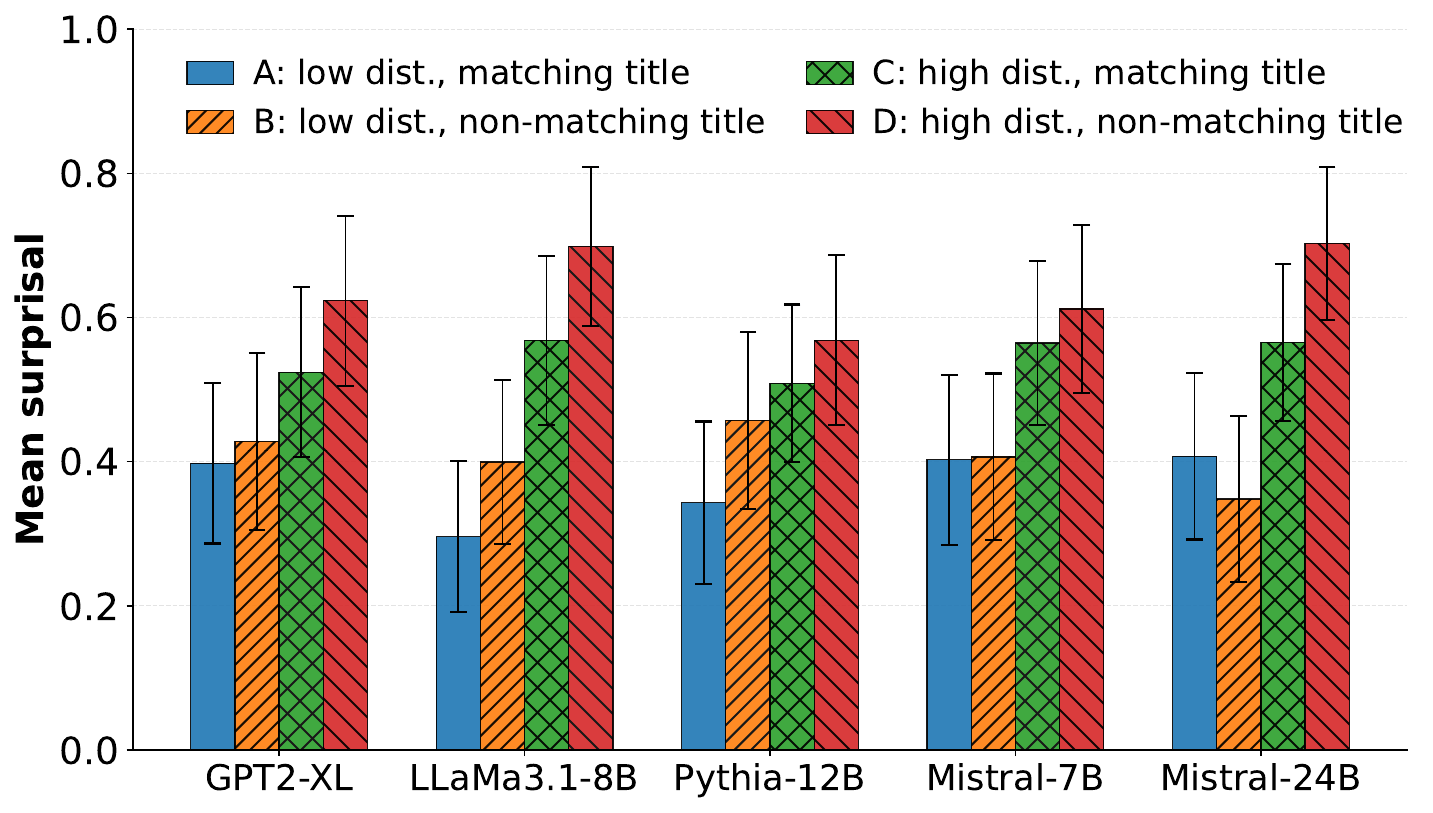}
\caption{Mean surprisal on the anaphor as a function of antecedent topicality and sentential distance (Exp. 1). The error bars are standard errors computed across the 16 texts.}
\label{fig_1_exp_1_surprisal}
\end{figure}

\begin{figure}[t]
\centering
\includegraphics[trim={0cm 0cm 0cm 0cm}, width=0.5\textwidth]{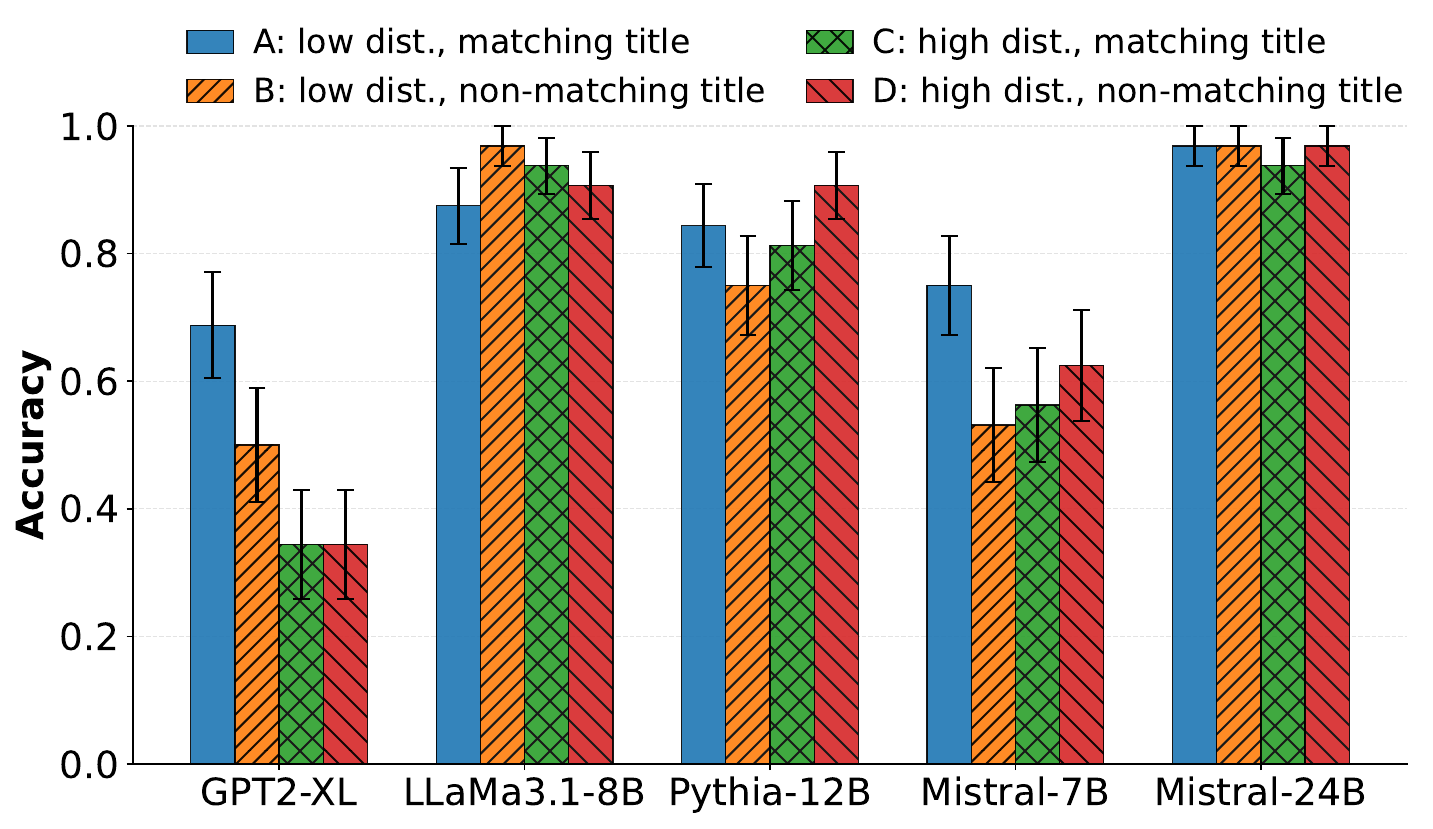}
\caption{Mean comprehension question accuracy as a function of antecedent topicality and sentential distance (Exp. 1). The error bars are standard errors.}
\label{fig_2_exp_1_accuracy}
\end{figure}



\subsection{Research Questions}
We evaluate whether LLMs are sensitive to the six factors delineated above that have been shown to affect anaphor resolution in humans. In particular, we test whether resolution is facilitated when antecedents are more accessible, due to greater discourse prominence, shorter sentential, spatial, or temporal distance, higher semantic similarity, and reduced interference from competing referents.

We address this question using five open-weight LLMs (GPT-2-XL, LLaMa-3.1-8B, Pythia-12B, Mistral-7B, and Mistral-24B). To assess process-level sensitivity, we adopt the standard linking hypothesis that relates model surprisal at the anaphor to human reading times \citep{hale2001a,levy2008expectation,wilcox2020on,li2024incremental}. To assess resolution success, we also measure model accuracy on comprehension questions that probe the antecedents of anaphors after text processing.

\section{Experiment 1}

Experiment 1 investigated the effect of (1) a thematic feature, whether the antecedent is topicalized by the text's title, and (2) a textual feature, the sentential distance (i.e., number of intervening sentences) between anaphors and antecedents, on anaphor resolution time and accuracy.

\subsection{Design and Materials}

The materials were from \citet{obrienej1987antecedent}. There were 16 texts, each occurring in four versions formed by orthogonally varying two factors. One factor was sentential distance (near, far), with the antecedent occurring either \textit{M} = 5.6 or \textit{M} = 15.2 sentences prior to the anaphor. The other factor was antecedent topicality (high, low), with the antecedent either focused by the title of the text or not. Thus, the four versions were:

\begin{itemize}[leftmargin=3em, labelsep=0.5em]
  \item [A.] near sentential distance, high antecedent topicality
  \item [B.] near sentential distance, low antecedent topicality
  \item [C.] far sentential distance, high antecedent topicality
  \item [D.] far sentential distance, low antecedent topicality
\end{itemize}

\subsection{Procedure and Dependent Measures}

\subsubsection{Anaphor Processing.} The surprisal of a probabilistic model in predicting the next token $i$ is $-\log_2(p_i)$ where $p_i$ is the probability of $i$. A common linking hypothesis is that the greater a model's surprisal for the next word, the longer the predicted reading time \citep{hale2001a,levy2008expectation,wilcox2020on,li2024incremental}. The surprisal for a text/version was computed as the average surprisal across the tokens making up the anaphor. Because these values were highly variable across the 16 texts, we normalized them within each text using:
$$\frac{surprisal-\min(surprisal_j)}{\max(surprisal_j) - \min(surprisal_j)}$$
\noindent where $surprisal_j$ denotes the set of the four surprisals for versions A-D. Thus, the normalized surprisals ranged from 0 to 1.  For each of the four versions A-D, we averaged the normalized surprisals across the 16 texts, resulting in four mean surprisals. We carried out this process for each LLM.

\subsubsection{Comprehension Question Answering.} For each of the 16 texts, there is a comprehension question asking for the antecedent of the anaphor. After an LLM processed a text, it was prompted with this question. We code the generated responses by adopting an "LLM-as-a-judge" paradigm using Gemini-2.5-flash-preview-09-2025 as the judge. Each question had a "gold" answer provided by us. The judge was prompted to first provide a brief justification comparing the candidate response to the gold answer and then output a binary accuracy label (1 = accurate, 0 = inaccurate). As with surprisal, for each of the four versions A-D, we averaged the accuracy score across the 16 texts, resulting in four mean accuracies. We repeated this process for each LLM.

\subsection{Results and Discussion}

\subsubsection{Surprisal Prediction of Reading Times.} Figure \ref{fig_1_exp_1_surprisal} shows, for each of the LLMs, the average surprisal for each of the four text versions. The prediction is that this should be lowest (i.e., that anaphor reading times should be fastest) when the title topicalizes the antecedent, leading readers to focus their attention on it during situation model construction, and when the sentential distance between the anaphor and antecedent is near. Thus, average surprisal should be lowest for version A, highest for version D, and intermediate for versions B and C. GPT-2-XL, Llama-3.1-8B, Pythia-12B, and Mistral-7B showed this pattern.

\subsubsection{Comprehension Question Accuracy.} Figure \ref{fig_2_exp_1_accuracy} shows, for each of the LLMs, the average accuracy for each of the four text versions. The predictions mirror those for surprisal, i.e., that accuracy should be highest when the title topicalizes the antecedent and when the sentential distance between the anaphor and antecedent is near. Thus, accuracy should be highest for version A, lowest for version D, and intermediate for the other versions B and C. GPT-2-XL showed this pattern. Among the other models, only Mistral-7B correctly ordered versions A and D. We note that because Llama-3.1-8B and Mistral-24B performed at ceiling on the comprehension questions, this might have limited our ability to detect differences between conditions.


\section{Experiment 2}
\begin{figure}[t]
\centering
\includegraphics[trim={0cm 0cm 0cm 0cm}, width=0.5\textwidth]{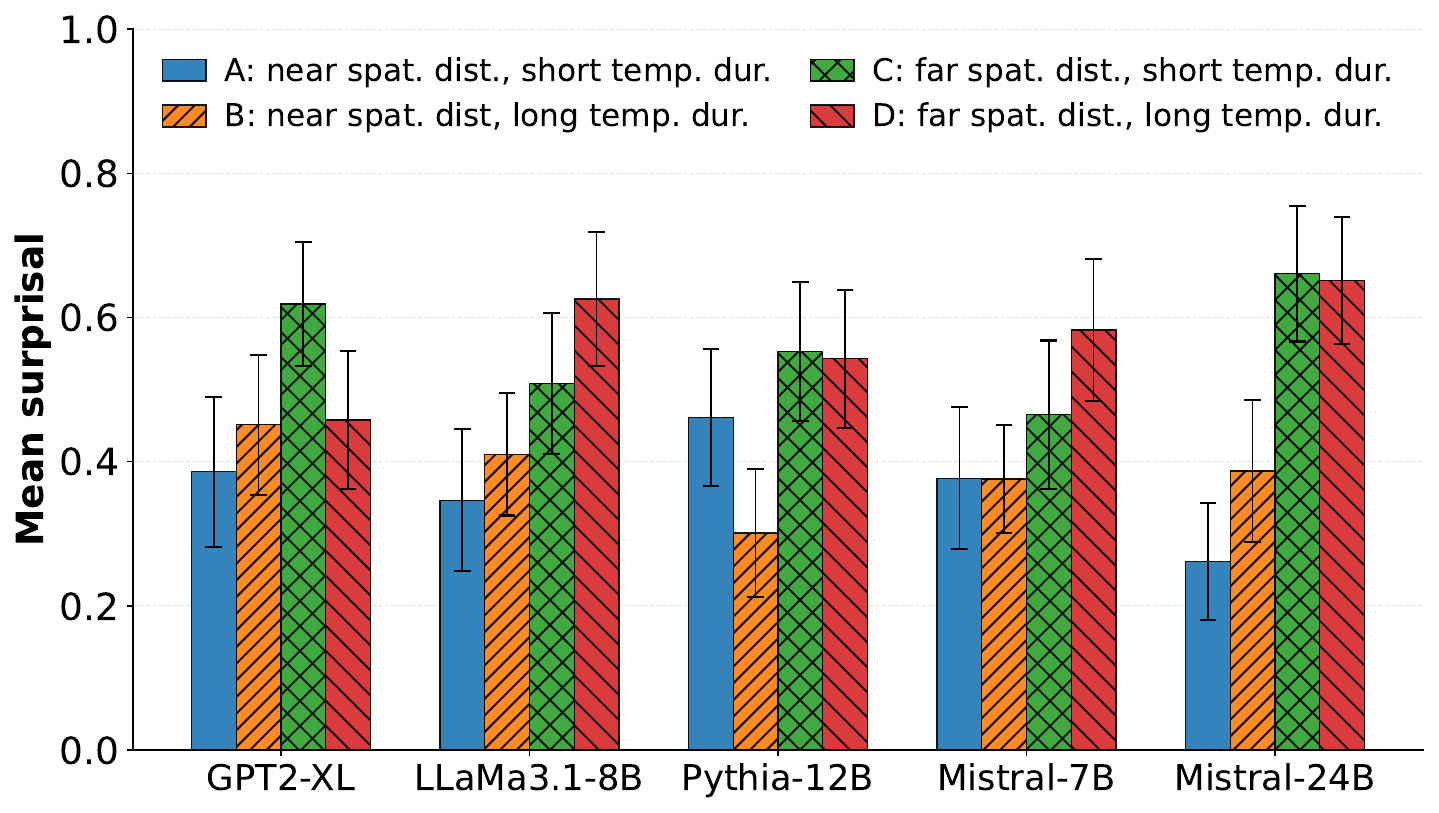}
\caption{Mean surprisal on the anaphor as a function of spatial distance and temporal duration (Exp. 2). The error bars are standard errors. }
\label{fig_3_exp_2_surprisal}
\end{figure}

\begin{figure}[t]
\centering
\includegraphics[trim={0cm 0cm 0cm 0cm}, width=0.5\textwidth]{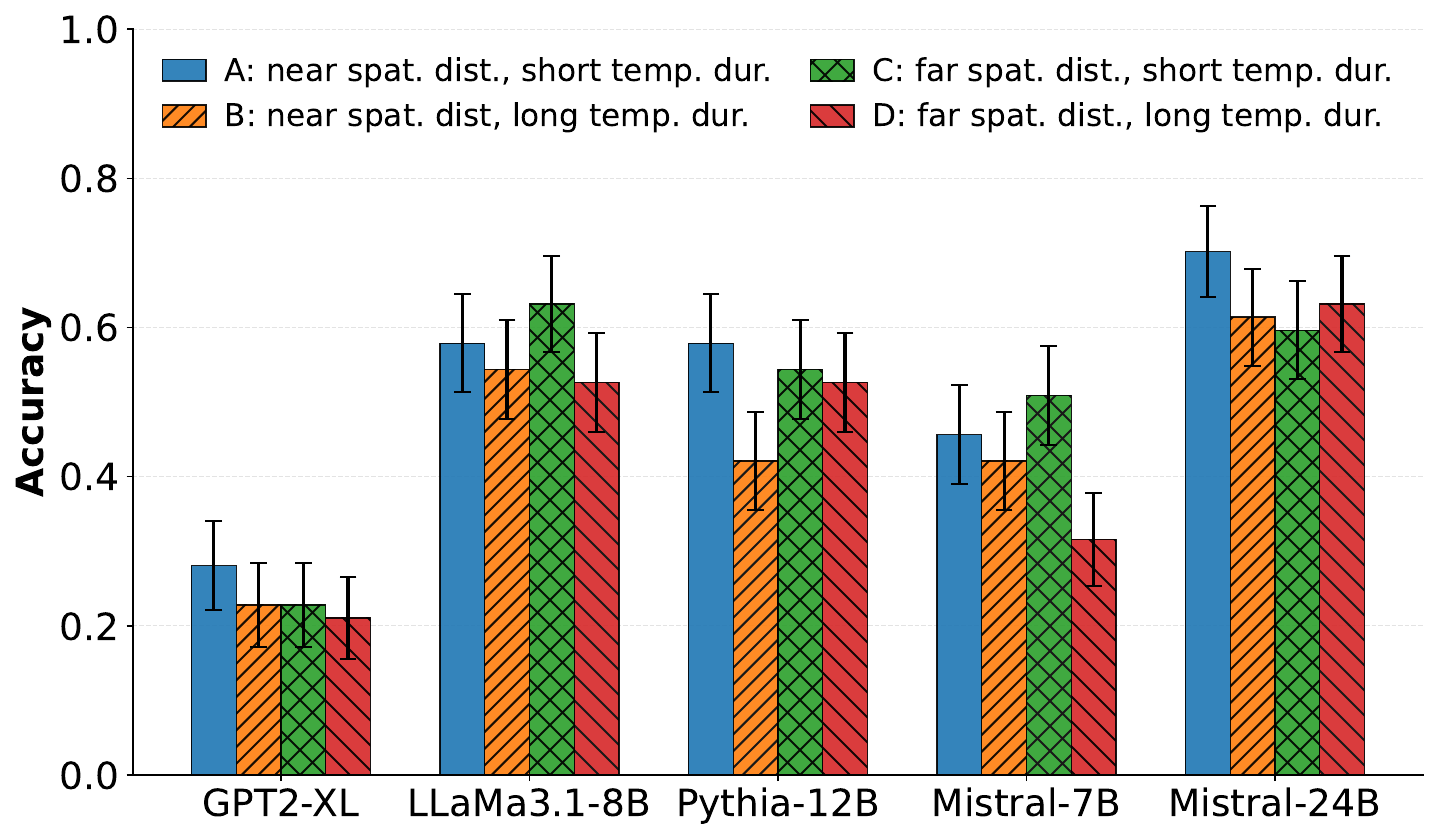}
\caption{Mean comprehension question accuracy as a function of spatial distance and temporal duration (Exp. 2). The error bars are standard errors.}
\label{fig_4_exp_2_accuracy}
\end{figure}
Experiment 2 investigated whether two contextual factors, spatial distance and temporal duration between anaphors and antecedents in readers' situation models, affect resolution time and accuracy. The details were the same as Experiment 1, except where otherwise noted.

\subsection{Design and Materials}

The materials were those of Experiment 1 of \citet{varma2019the}. There were 19 texts, and each occurred in four versions formed by orthogonally varying the spatial distance (near, far) and temporal duration (short, long) between the anaphor and antecedent in the reader's situation model:

\begin{itemize}[leftmargin=3em]
  \item [A.] near spatial distance, short temporal duration
  \item [B.] near spatial distance, long temporal duration
  \item [C.] far spatial distance, short temporal duration
  \item [D.] far spatial distance, long temporal duration
\end{itemize}

\noindent These factors were manipulated in the \textit{M} = 15.1 (\textit{SD} = 2.4) sentences that separated anaphors from their antecedents. The same categorical anaphor (e.g., \textit{metal furniture})  and typical antecedent (e.g., \textit{metal chair}) were used for each of the four versions A-D.

\subsection{Results and Discussion}

\subsubsection{Surprisal Prediction of Reading Times.} Figure \ref{fig_3_exp_2_surprisal} shows the average surprisal of each of the five models on each of the four text versions. The prediction is that this should be lowest (i.e., anaphor resolution fastest) for version A, highest (i.e., anaphor resolution slowest) for version D, and intermediate for versions B and C. Llama-3.1-8B and Mistral-7B show the predicted pattern.

\subsubsection{Comprehension Question Accuracy.} Figure \ref{fig_4_exp_2_accuracy} shows the average accuracies. The predictions parallel those for surprisal: the highest average accuracy is expected for version A, the lowest for version D, and intermediate values for versions B and C. Only GPT-2-XL shows this pattern. That said, all of the other models correctly order versions A and D.

\section{Experiment 3}
\begin{figure}[t]
\centering
\includegraphics[trim={0cm 0cm 0cm 0cm}, width=0.5\textwidth]{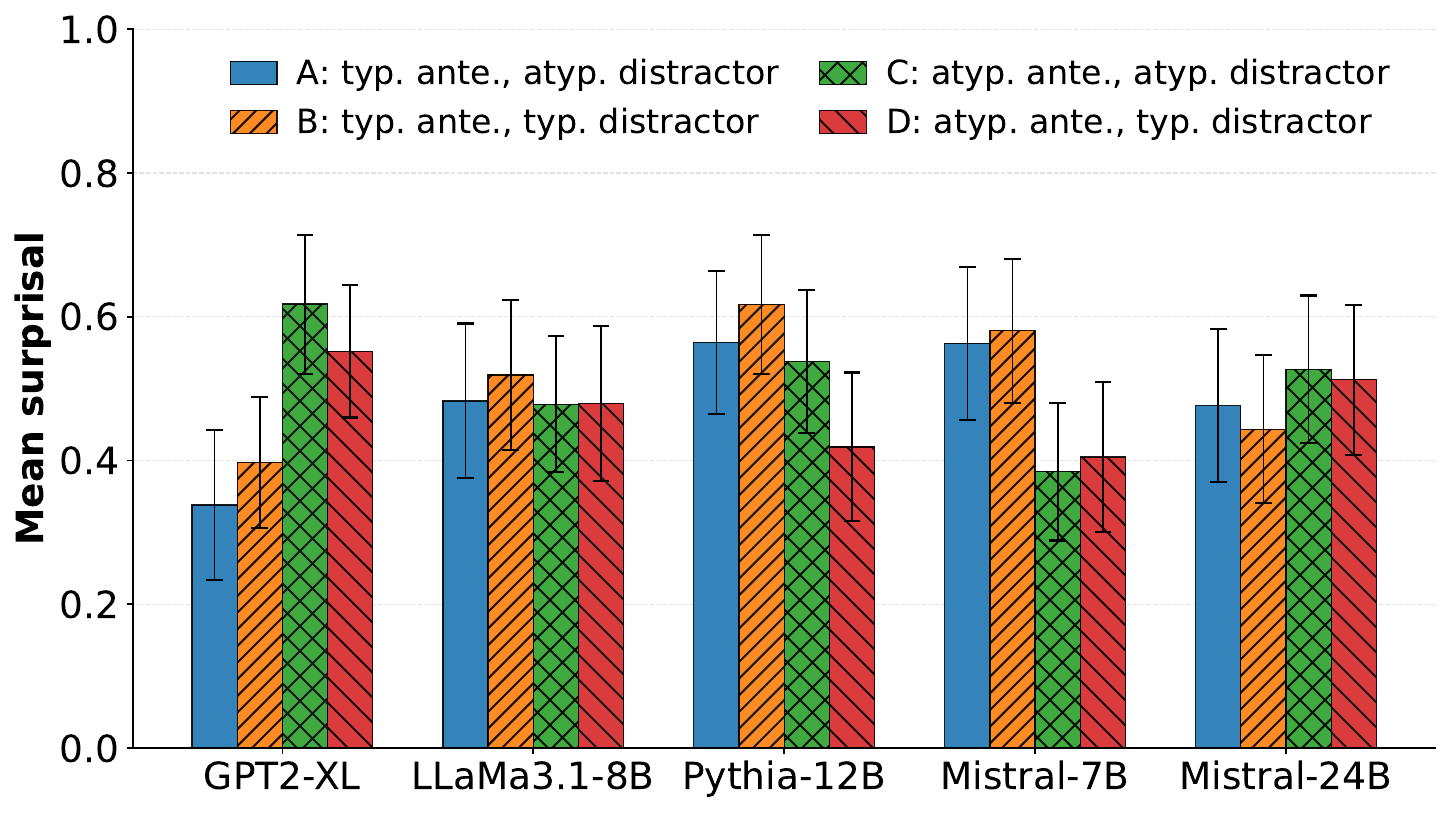}
\caption{Mean surprisal on the anaphor as a function of the semantic overlap and semantic interference factors (Exp. 3). The error bars are standard errors.}
\label{fig_5_exp_3_surprisal}
\end{figure}

\begin{figure}[t]
\centering
\includegraphics[trim={0cm 0cm 0cm 0cm}, width=0.5\textwidth]{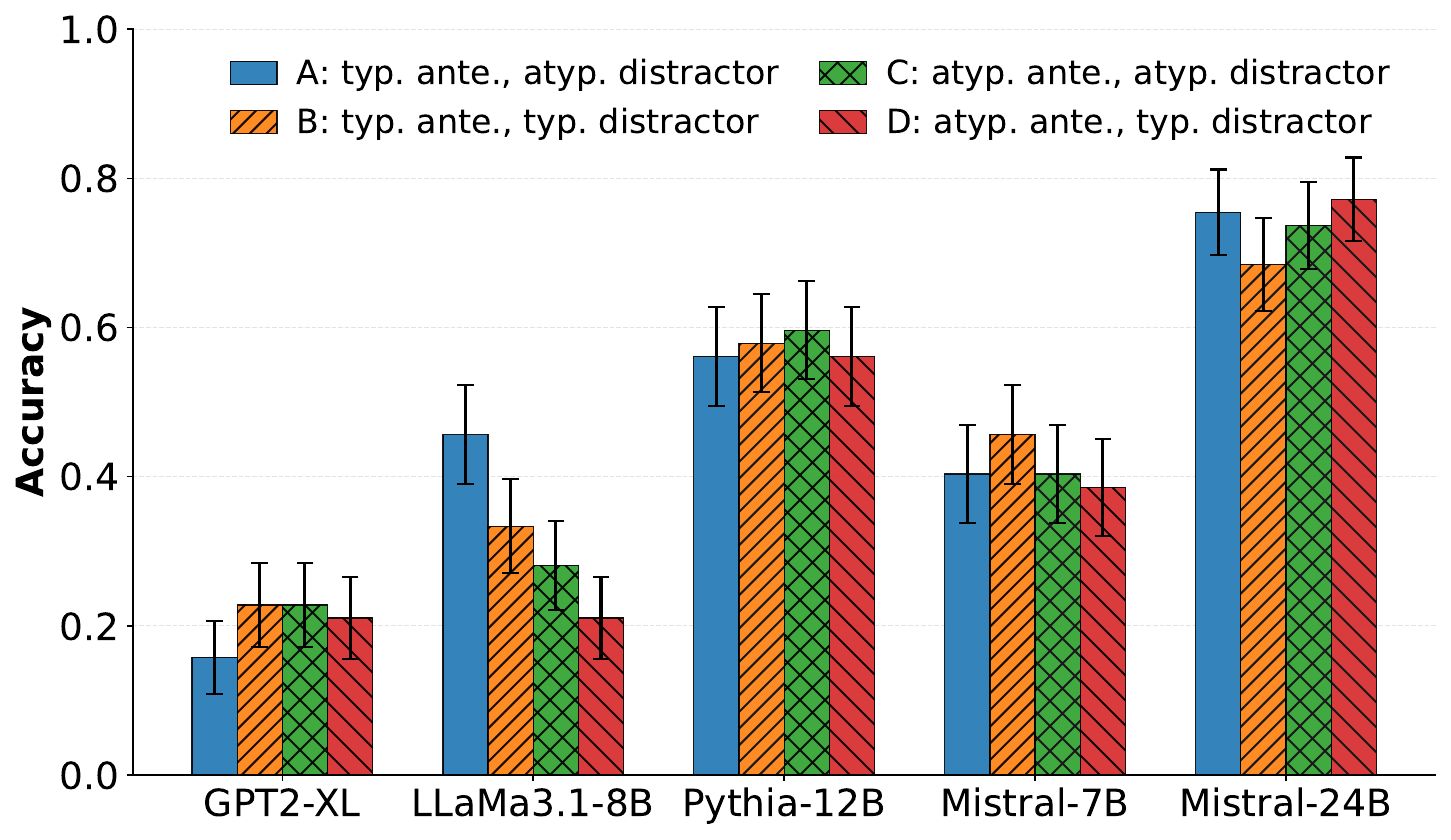}
\caption{Mean comprehension question accuracy as a function of the semantic overlap and semantic interference factors (Exp. 3). The error bars are standard errors.}
\label{fig_6_exp_3_accuracy}
\end{figure}
Experiment 3 investigated whether two semantic factors, the semantic overlap between the anaphor and antecedent and the semantic interference from non-antecedent distractors, affect resolution time and accuracy. Except where otherwise noted, the details are the same as previous experiments.

\subsection{Design and Materials}

The materials were those of Experiment 2 of \citet{varma2019the}: 19 texts, each occurring in four versions formed by orthogonally varying the semantic overlap between the antecedent and anaphor (high, low) and the semantic interference caused by a non-antecedent distractor (low, high). The anaphors were categorical (e.g., \textit{metal furniture}). Following prior work \citep{corbett1984prenominal,garrod1977interpreting,varma2019the}, high semantic overlap was operationalized by antecedents that were typical members of the category (e.g., \textit{chair}) and low semantic overlap by antecedents that were atypical (e.g., \textit{stool}). Low semantic interference was operationalized by non-antecedent distractors that are atypical members and high semantic interference by non-antecedent distractors that are typical. Thus, the four versions were:

\begin{itemize}[leftmargin=3em, labelsep=0.5em]
  \item [A.] typical antecedent, atypical distractor
  \item [B.] typical antecedent, typical distractor
  \item [C.] atypical antecedent, atypical distractor
  \item [D.] atypical antecedent, typical distractor
\end{itemize}

\noindent The base texts were the version D (far spatial distance, long temporal duration) texts from Experiment 2. The four versions of each text shared the same categorical anaphor. Only the antecedent and the non-antecedent distractors were varied.

\subsection{Results and Discussion}

\subsubsection{Surprisal Prediction of Reading Times.} The average surprisal of the five models on the four text versions is shown in Figure \ref{fig_5_exp_3_surprisal}. The prediction is that the anaphor resolution is fastest for version A, slowest for D, and intermediate for versions B and C. None of the models show the predicted pattern, although GPT-2-XL correctly orders versions A and D.

\subsubsection{Comprehension Question Accuracy.} The average accuracy of the five models on the four text versions is shown in Figure \ref{fig_6_exp_3_accuracy}. The predictions mirror those for surprisal on the anaphor, with the highest average accuracy expected for version A, the lowest for version D, and intermediate values for versions B and C. Only Llama-3.1-8B shows the predicted pattern, although its overall accuracies are quite low.

\section{General Discussion}

\begin{table}[t]
\centering
\setlength{\tabcolsep}{2pt}
\renewcommand{\arraystretch}{1.05}

\resizebox{\columnwidth}{!}{
\begin{tabular}{l p{3cm} p{3cm}}
\hline
 & \multicolumn{2}{c}{\textbf{Measures}} \\
\textbf{Factor} & \multicolumn{1}{c}{\textbf{Surprisal}} & \multicolumn{1}{c}{\textbf{Accuracy}} \\
\hline


\parbox[c][3\baselineskip][c]{4.4cm}{\raggedright Topicality (Exp.\ 1)}
&
\parbox[c][3\baselineskip][c]{3cm}{
\begin{tabular}{@{}l@{}}
All models
\end{tabular}}
&
\parbox[c][3\baselineskip][c]{3cm}{
\begin{tabular}{@{}l@{}}
GPT2-XL\\
\end{tabular}}
\\
\hline

\parbox[c][3\baselineskip][c]{4.4cm}{\raggedright Sentential distance (Exp.\ 1)}
&
\parbox[c][3\baselineskip][c]{3cm}{
\begin{tabular}{@{}l@{}}
All models
\end{tabular}}
&
\parbox[c][3\baselineskip][c]{3cm}{
\begin{tabular}{@{}l@{}}
GPT2-XL\\
(others partial)\\
\end{tabular}}
\\
\hline

\parbox[c][3\baselineskip][c]{4.4cm}{\raggedright Spatial distance (Exp.\ 2)}
&
\parbox[c][3\baselineskip][c]{3cm}{
\begin{tabular}{@{}l@{}}
All models \\ except Pythia-12B
\end{tabular}}
&
\parbox[c][3\baselineskip][c]{3cm}{
\begin{tabular}{@{}l@{}}
None/weak effects\\
\end{tabular}}
\\
\hline

\parbox[c][3\baselineskip][c]{4.4cm}{\raggedright Temporal duration (Exp.\ 2)}
&
\parbox[c][3\baselineskip][c]{3cm}{
\begin{tabular}{@{}l@{}}
LLaMa-3.1-8B\\
Mistral-7B \\
\end{tabular}}
&
\parbox[c][3\baselineskip][c]{3cm}{
\begin{tabular}{@{}l@{}}
None/weak effects\\
\end{tabular}}
\\
\hline

\parbox[c][3\baselineskip][c]{4.4cm}{\raggedright Antecedent similarity (Exp.\ 3)}
&
\parbox[c][3\baselineskip][c]{3cm}{
\begin{tabular}{@{}l@{}}
GPT2-XL\\
\end{tabular}}
&
\parbox[c][3\baselineskip][c]{3cm}{
\begin{tabular}{@{}l@{}}
LLaMa-3.1-8B\\
(low overall accuracy)\\
\end{tabular}}
\\
\hline

\parbox[c][3\baselineskip][c]{4.4cm}{\raggedright Semantic interference (Exp.\ 3)}
&
\parbox[c][3\baselineskip][c]{3cm}{
\begin{tabular}{@{}l@{}}
GPT2-XL\\
\end{tabular}}
&
\parbox[c][3\baselineskip][c]{3cm}{
\begin{tabular}{@{}l@{}}
None/weak effects\\
\end{tabular}}
\\
\hline

\end{tabular}
}
\caption{Summary of where LLMs show human-like directional patterns in anaphor resolution across discourse, situation-model, and semantic factors.}
\end{table}

Cognitive science studies have identified factors that affect the speed and success of anaphor resolution. This study examined whether the same factors affect anaphor resolution in LLMs. Positive results would constitute evidence that LLMs can serve as cognitive models of human anaphor resolution, and perhaps of language understanding more generally.

Experiment 1 investigated the effects of (1) antecedent topicality and (2) sentential distance. The remaining experiments focused on the reader's situation model of the text. Experiment 2 investigated the contextual effects of (3) spatial distance and (4) temporal duration; Experiment 3 investigated the semantic effects of (5) antecedent similarity and (6) interference from non-antecedent distractors. Two indices of model performance, surprisal on the anaphor and accuracy on a comprehension question about the antecedent, were compared with the corresponding human measures of reading time and accuracy. The models considered were GPT-2-XL, Llama-3.1-8B, Pythia, Mistral-24B, and Mistral-7B. To varying degrees, the models showed the six effects documented in the literature, whether in their surprisal values or comprehension accuracies. The model with the greatest overall alignment to human anaphor resolution was perhaps Mistral-7B, although it failed to account for the effects of (5) antecedent similarity and (6) interference in Experiment 3. GPT-2-XL achieved surprisingly respectable overall alignment given that it is the oldest and smallest of the models.

\emph{A notable pattern across experiments is that LLMs more reliably exhibited human-like sensitivity to discourse prominence and distance-based factors than to semantic similarity and interference.} This asymmetry is informative. 
Effects of sentential, spatial, and temporal distance reflect graded accessibility, which may be approximated by recency biases and attention dynamics in sequential language models. In contrast, semantic interference effects require structured competition among partially overlapping representations during memory retrieval, a process central to cognitive theories of anaphor resolution but not explicitly implemented in current LLM architectures. From this perspective, partial alignment does not undermine the present findings; rather, it helps localize where LLMs diverge from human discourse processing.

Thus, we have some evidence of cognitive alignment between LLMs, especially Mistral-7B and GPT-2-XL, and human performance, suggesting their potential utility as cognitive science models. Still, work remains to be done. For example, although the profiles of these models' comprehension question accuracies across passage versions were generally human-like, their absolute performance was quite low.

The largest limitation of the current study is the lack of items (i.e., texts), which precluded running statistical analyses to better establish the informal trends present in the model results. Here, we were limited by the relatively small size of cognitive science studies relative to ML studies and the general unavailability of materials and human datasets for classic studies of anaphor resolution. It would have been easy to generate multiple 'participants' by increasing the temperature parameter, running each model multiple times, and computing statistics across these samples. However, we question the validity of this approach. There is no reason to believe that such simulated "participants" vary in the same ways that humans do: in their working memory capacity, executive function, domain knowledge, reading skill, etc. For this reason, we limited ourselves to providing descriptive, summary statistics and cautiously interpreting the observed trends. We also recognize the limitations of our LLM-as-a-judge technique for scoring comprehension accuracy. Although it provides a reasonable heuristic, the binary scoring and reliance on an automatic judge introduce known biases and can be different from human judgments. It is best to treat the reported accuracies as approximations.

One goal for future research is to examine whether other factors that affect anaphor resolution performance in humans also affect LLMs. Some of these are relatively subtle, such as whether an antecedent and anaphor belong to the same event within a narrative text or whether they are in different events separated by a boundary \citep{thompson2016event}. Whether or not LLMs are sensitive to such factors is an important question for cognitive scientists evaluating their potential as models of human language understanding. Examination of a broad range of factors known to affect human anaphor resolution might also guide NLP researchers as they design the new benchmarks for measuring the coreference resolution abilities of LLMs \citep{gan2023assessing,manikantan2024identifyme}.





\printbibliography

@article{anderson1983the,
  author = {Anderson, A. and Garrod, S. C. and Sanford, A. J.},
  title = {The accessibility of pronominal antecedents as a function of episode shifts in narrative text},
  year = {1983},
  journaltitle = {The Quarterly Journal of Experimental Psychology A},
  volume = {35},
  pages = {427--440},
}

@incollection{cambria2025semantics,
  author = {Cambria, E.},
  title = {Semantics Processing},
  year = {2025},
  publisher = {Springer},
}

@article{clark1979in,
  author = {Clark, H. H. and Sengul, C. J.},
  title = {In search of referents for nouns and pronouns},
  year = {1979},
  journaltitle = {Memory \& Cognition},
  volume = {7},
  pages = {35--41},
  doi = {10.3758/BF03196932},
}

@article{corbett1984prenominal,
  author = {Corbett, A. T.},
  title = {Prenominal adjectives and the disambiguation of anaphoric nouns},
  year = {1984},
  journaltitle = {Journal of Verbal Learning and Verbal Behavior},
  volume = {23},
  pages = {683--695},
}

@article{daneman1980individual,
  author = {Daneman, M. and Carpenter, P. A.},
  title = {Individual differences in working memory and reading},
  year = {1980},
  journaltitle = {Journal of Verbal Learning and Verbal Behavior},
  volume = {19},
  pages = {450--466},
}

@article{devlin2019bert,
  author = {Devlin, J. and Chang, M.-W. and Lee, K. and Toutanova, K.},
  title = {BERT: Pre-training of deep bidirectional transformers for language understanding},
  year = {2019},
}

@inproceedings{gan2023assessing,
    title = "Assessing the Capabilities of Large Language Models in Coreference: An Evaluation",
    author = "Gan, Yujian  and
      Poesio, Massimo  and
      Yu, Juntao",
    editor = "Calzolari, Nicoletta  and
      Kan, Min-Yen  and
      Hoste, Veronique  and
      Lenci, Alessandro  and
      Sakti, Sakriani  and
      Xue, Nianwen",
    booktitle = "Proceedings of the 2024 Joint International Conference on Computational Linguistics, Language Resources and Evaluation (LREC-COLING 2024)",
    month = may,
    year = "2024",
    address = "Torino, Italia",
    publisher = "ELRA and ICCL",
    url = "https://aclanthology.org/2024.lrec-main.145/",
    pages = "1645--1665"
}

@article{garrod1977interpreting,
  author = {Garrod, S. and Sanford, A. J.},
  title = {Interpreting anaphoric relations: the integration of semantic information while reading},
  year = {1977},
  journaltitle = {Journal of Verbal Learning and Verbal Behavior},
  volume = {16},
  pages = {77--90},
}

@misc{grattafiori2024the,
  author = {Grattafiori, A. and others},
  title = {The Llama 3 Herd of Models},
  year = {2024},
  howpublished = {arXiv},
  url = {https://doi.org/10.48550/arXiv.2407.21783},
}

@inproceedings{hale2001a,
  author = {Hale, J.},
  title = {A probabilistic Earley parser as a psycholinguistic model},
  year = {2001},
  booktitle = {Proceedings of the 2nd Meeting of NAACL},
}

@article{ivanova2025how,
  author = {Ivanova, A. A.},
  title = {How to evaluate the cognitive abilities of LLMs},
  year = {2025},
  journaltitle = {Nature Human Behavior},
}

@article{levy2008expectation,
  author = {Levy, R.},
  title = {Expectation-based syntactic comprehension},
  year = {2008},
  journaltitle = {Cognition},
  volume = {106},
  pages = {1126--1177},
}

@inproceedings{li2024incremental,
  author = {Li, A. and Cai, T. and Feng, X. and Narang, S. and Peng, A. and Shah, R. S. and Varma, S.},
  title = {Incremental comprehension of garden-path sentences by Large Language Models: Semantic interpretation, syntactic re-analysis, and attention},
  year = {2024},
  booktitle = {Proceedings of the 46th Annual Conference of the Cognitive Science Society},
  pages = {6069--6076},
}

@misc{liu2025enhancing,
  author = {Liu, X. and Deng, S. and Wang, M. and Dong, Z. and Dai, L. and Li, J. and Nong, R.},
  title = {Enhancing Coreference Resolution with Pretrained Language Models: Bridging the Gap Between Syntax and Semantics},
  year = {2025},
  howpublished = {arXiv},
  url = {https://doi.org/10.48550/ARXIV.2504.05855},
}

@misc{manikantan2024identifyme,
  author = {Manikantan, K. and Tapaswi, M. and Gandhi, V. and Toshniwal, S.},
  title = {IdentifyMe: A challenging long-context mention resolution benchmark},
  year = {2024},
  howpublished = {arXiv},
}

@misc{mcnamara2009misc,
  author = {McNamara, D. S. and Magliano, J.},
  year = {2009},
  title = {Toward a comprehensive model of comprehension},
  note = {In B. Ross (Ed.), The psychology of learning and motivation},
}

@article{morrow1987accessibility,
  author = {Morrow, D. G. and Greenspan, S. L. and Bower, G. H.},
  title = {Accessibility and situation models in narrative comprehension},
  year = {1987},
  journaltitle = {Journal of Memory and Language},
  volume = {26},
  pages = {165--187},
}

@inproceedings{novk2025findings,
  author = {Novák, M. and others},
  title = {Findings of the Fourth Shared Task on Multilingual Coreference Resolution},
  year = {2025},
  booktitle = {Proceedings of the CRAC Shared Task},
  doi = {10.18653/v1/2025.crac-1.9},
}

@article{obrienej1987antecedent,
  author = {O'Brien, E. J.},
  title = {Antecedent search processes and the structure of text},
  year = {1987},
  journaltitle = {Journal of Experimental Psychology: Learning, Memory, and Cognition},
  volume = {13},
  pages = {278--290},
}

@article{piantadosi2024why,
  author = {Piantadosi, S. T. and others},
  title = {Why concepts are (probably) vectors},
  year = {2024},
  journaltitle = {Trends in Cognitive Sciences},
  volume = {28},
  pages = {844--856},
}

@article{radford2019language,
  author = {Radford, A. and Wu, J. and Child, R. and Luan, D. and Amodei, D. and Sutskever, I.},
  title = {Language models are unsupervised multitask learners},
  year = {2019},
}

@article{rinck1995anaphora,
  author = {Rinck, M. and Bower, G. H.},
  title = {Anaphora resolution and the focus of attention in situation models},
  year = {1995},
  journaltitle = {Journal of Memory and Language},
  volume = {34},
  pages = {110--131},
}

@article{rosch1976basic,
  author = {Rosch, E. and Mervis, C. B. and Gray, W. D. and Johnson, D. M. and Boyes-Braem, P.},
  title = {Basic objects and natural categories},
  year = {1976},
  journaltitle = {Cognitive Psychology},
  volume = {9},
  pages = {382--440},
}

@misc{shah2025the,
  author = {Shah, R. S. and Varma, S.},
  title = {The potential and the pitfalls of using pre-trained language models as cognitive science theories},
  year = {2025},
  howpublished = {arXiv},
  url = {https://arxiv.org/abs/2501.12651},
}

@inproceedings{talukdar2025coreference,
  author = {Talukdar, C. and Rahman, M.},
  title = {Coreference Resolution in Machine Learning: A Survey},
  year = {2025},
  booktitle = {2025 IEEE Guwahati Subsection Conference (GCON)},
  doi = {10.1109/GCON65540.2025.11173320},
}

@article{thompson2016event,
  author = {Thompson, A. N. and Radvansky, G. A.},
  title = {Event boundaries and anaphoric reference},
  year = {2016},
  journaltitle = {Psychonomic Bulletin \& Review},
  volume = {23},
  pages = {849--856},
}

@book{vandijk1983strategies,
  author = {van Dijk, T. A. and Kintsch, W.},
  title = {Strategies of discourse comprehension},
  year = {1983},
  publisher = {Academic Press},
}

@article{varma2019the,
  author = {Varma, S. and Janssen, A.},
  title = {The structure of situation models as revealed by anaphor resolution},
  year = {2019},
  journaltitle = {Language Sciences},
  volume = {72},
  pages = {104--115},
}

@misc{wang2024mmlu,
  author = {Wang, Y. and others},
  title = {MMLU-Pro: A More Robust and Challenging Multi-Task Language Understanding Benchmark},
  year = {2024},
  howpublished = {arXiv},
  url = {https://doi.org/10.48550/arXiv.2406.01574},
}

@inproceedings{wilcox2020on,
  author = {Wilcox, E. and Gauthier, J. and Hu, J. and Qian, P. and Levy, R. P.},
  title = {On the predictive power of neural language models for human real-time comprehension behavior},
  year = {2020},
  booktitle = {Proceedings of the 42nd Annual Meeting of the Cognitive Science Society},
  pages = {1707--1713},
}

@article{zwaan1996processing,
  author = {Zwaan, R. A.},
  title = {Processing narrative time shifts},
  year = {1996},
  journaltitle = {Journal of Experimental Psychology: Learning, Memory, and Cognition},
  volume = {22},
  pages = {1196--1207},
}

@misc{openai_gpt5_2025,
  title        = {Introducing GPT-5},
  author       = {{OpenAI}},
  year         = {2025},
  month        = Aug,
  url          = {https://openai.com/index/introducing-gpt-5/},
  note         = {Accessed: 2026-01-30}
}

@misc{mistral_small3_2025,
  title        = {Mistral Small 3},
  author       = {{Mistral AI Team}},
  year         = {2025},
  month        = jan,
  day          = {30},
  url          = {https://mistral.ai/news/mistral-small-3},
  note         = {Accessed: 2026-01-30}
}

@inproceedings{charpentier-etal-2025-findings,
    title = "Findings of the Third {B}aby{LM} Challenge: Accelerating Language Modeling Research with Cognitively Plausible Data",
    author = {{BabyLM Organizers}},
    editor = "Charpentier, Lucas  and
      Choshen, Leshem  and
      Cotterell, Ryan  and
      Gul, Mustafa Omer  and
      Hu, Michael Y.  and
      Liu, Jing  and
      Jumelet, Jaap  and
      Linzen, Tal  and
      Mueller, Aaron  and
      Ross, Candace  and
      Shah, Raj Sanjay  and
      Warstadt, Alex  and
      Wilcox, Ethan Gotlieb  and
      Williams, Adina",
    booktitle = "Proceedings of the First BabyLM Workshop",
    month = nov,
    year = "2025",
    address = "Suzhou, China",
    publisher = "Association for Computational Linguistics",
    url = "https://aclanthology.org/2025.babylm-main.28/",
    doi = "10.18653/v1/2025.babylm-main.28",
    pages = "399--420"
}

@inproceedings{hu-etal-2020-closer,
    title = "A closer look at the performance of neural language models on reflexive anaphor licensing",
    author = "Hu, Jennifer  and
      Chen, Sherry Yong  and
      Levy, Roger",
    booktitle = "Proceedings of the Society for Computation in Linguistics 2020",
    month = jan,
    year = "2020",
    address = "New York, New York",
    publisher = "Association for Computational Linguistics",
    url = "https://aclanthology.org/2020.scil-1.39/",
    pages = "323--333"
}

@inproceedings{lee-schuster-2022-language,
    title = "Can language models capture syntactic associations without surface cues? A case study of reflexive anaphor licensing in {E}nglish control constructions",
    author = "Lee, Soo-Hwan  and
      Schuster, Sebastian",
    booktitle = "Proceedings of the Society for Computation in Linguistics 2022",
    month = feb,
    year = "2022",
    address = "online",
    publisher = "Association for Computational Linguistics",
    url = "https://aclanthology.org/2022.scil-1.18/",
    pages = "206--211"
}

@inproceedings{pandit-hou-2021-probing,
    title = "Probing for Bridging Inference in Transformer Language Models",
    author = "Pandit, Onkar  and
      Hou, Yufang",
    booktitle = "Proceedings of the 2021 Conference of the North American Chapter of the Association for Computational Linguistics: Human Language Technologies",
    month = jun,
    year = "2021",
    address = "Online",
    publisher = "Association for Computational Linguistics",
    url = "https://aclanthology.org/2021.naacl-main.327/",
    doi = "10.18653/v1/2021.naacl-main.327",
    pages = "4153--4163"
}

\end{document}